\documentclass{article}
\usepackage{ijcai20}
\usepackage{subfigure}
\usepackage{times}
\usepackage{soul}
\usepackage{url}
\usepackage[utf8]{inputenc}
\usepackage[small]{caption}
\usepackage{graphicx}
\usepackage{amsmath}
\usepackage{amsthm}
\usepackage{booktabs}
\usepackage{algorithm}
\usepackage{algorithmic}
\usepackage{amssymb} 
\usepackage{latexsym}
\usepackage{xspace}
\usepackage{enumerate}
\usepackage[defblank]{paralist}
\usepackage{microtype}
\usepackage{comment}
\usepackage{todonotes}
\usepackage{tabularx}
\usepackage{booktabs}
\usepackage{multirow}
\usepackage{MnSymbol}
\usepackage{nameref}

\newcommand{\LDLf}{{\sc ldl}$_f$\xspace}
\newcommand{\commentout}[1]{}

\title{Learning and Solving Regular Decision Processes}

\author{
Eden Abadi
\And
Ronen I. Brafman
\affiliations
Department of Computer Science, Ben Gurion University, Israel\\
\emails
\
abadied@post.bgu.ac.il,
brafman@cs.bgu.ac.il
}

\begin{document}

\maketitle

\begin{abstract}
Regular Decision Processes (RDPs) are a recently introduced model that extends MDPs with non-Markovian
dynamics and rewards. The non-Markovian behavior is restricted to depend on regular properties of the history.
These can be specified using regular expressions or formulas in linear dynamic logic over finite traces. 
Fully specified RDPs can be solved by compiling them into an appropriate MDP.
Learning RDPs from data is a challenging problem that has yet to be addressed, on which
we focus in this paper. Our approach rests on a new representation for
RDPs using Mealy Machines that emit a distribution and an expected reward for each state-action pair.
Building on this representation, we combine automata learning techniques with history clustering to learn such
a Mealy machine and solve it by adapting MCTS to it. We empirically evaluate this approach, demonstrating its
feasibility.
\end{abstract}

\section{Introduction}
In the emerging area of personal health tracking, one records one's pulse, blood pressure, glucose levels, activity levels, nutritional information and much more, in an attempt to learn how to improve one's physical and mental health.
In this domain, the state of many variables of interest and the effects of various actions are most likely {\em not} Markovian functions of the value of the most recently measured variables. Hence, applying standard, MDP-based, RL algorithms~\cite{Sutton2} to a state model consisting of the value of these observed variables  will most likely lead to sub-optimal behavior. 

Motivated by such application domains, Regular Decision Processes (RDPs)~\cite{BrafmanG19} have been recently introduced as a non-Markovian extension of MDPs that does not require knowing or hypothesizing a hidden state. 
An RDP is a fully observable, non-Markovian model in which the next state and reward are a stochastic function of the {\em entire} history of the system. 
However, this dependence on the past is restricted to regular functions. That is, the next state distribution and reward depends on which {\em regular} expression the history satisfies. 

An RDP can be transformed into an MDP by extending the state of the RDP with variables that track the satisfaction of
the regular expression governing the RDP dynamics. Thus, essentially, to learn an RDP, we need to learn these regular
expressions. 
For example,  in the context of personal health-tracking, one might learn which sequences of activities and measurements make a state of hyperglycaemia, hypoglycaemia, or depression, likely, and consequently,
adapt behavior policies that prevent them.  

An optimal policy for an RDP is a mapping from regular properties of history to actions. Thus, it provides users with clear, understandable guidelines, based on {\em observable}  properties of the world, in contrast to, e.g., arbitrary hidden states in a learned POMDP, or 
 unclear features in a neural network.

This paper makes two contributions to the emerging theory of RDPs.
Our first contribution is the use of a deterministic Mealy Machine to specify the RDP.
For each state and action, this Mealy machine emits as its output, a class label. This class label is associated with
a distribution over the underlying system states and a reward signal.
This idea extends the use of Mealy Machines to specify non-Markovian rewards, introduced  recently by~\cite{RewardMachines},
to RDPs.
Our second, and main contribution is to use this idea to formulate the first algorithm for learning RDPs from data,
and to evaluate it on two non-Markovian domains.  
Our algorithm  identifies, through exploration, histories that have similar dynamics based on their empirical distributions 
and then learns a Mealy Machine that outputs, for each history, an appropriate label.
Then, we solve this Mealy Machine to obtain an optimal policy. This can be done by either reducing it to an
explicit MDP, or by using the Mealy Machine to provide the distributions needed for running MCTS~\cite{KocsisS06,SilverV10}. 
This algorithm was implemented and tested on two domains modelled as RDPs, demonstrating its ability to learn RDPs from observable data
and to generate a near-optimal policy for these models.

\section{Background}
We assume familiarity with MDPs, recalling basic notation only. We briefly discuss
NMDPS, RDPs, and Mealy Machine.

\subsection{MDP and NMDPs}

A Markov Decision Process (MDP) is a tuple $M = \langle S,A,Tr,R, s_0\rangle$. $S$ is the set of states, $A$ a set of actions, $Tr: S \times A \rightarrow \Pi(S)$ is the transition function that returns for every state $s$ and action $a$ the distribution  over the next state. $R: S \times A \rightarrow \mathcal{R}$ is the reward function that returns the real valued reward received by the agent after performing action $a$ at state $s$, and $s_0 \in S$ is the initial state.

A solution to an MDP is a {\em policy} $\rho: S\rightarrow A$ that maps each state to an action.
The {\em value} of  $\rho$ at state $s$, $v^\rho (s)$, is the expected sum of discounted rewards when starting at state $s$ and selecting actions based on $\rho$. An {\em optimal}  policy, denoted $\rho^*$, maximizes
the expected  sum of discounted rewards for every starting state $s\in S$. Such a policy always exists for infinite horizon discounted reward problems.

A non-Markovian Decision Process (NMDP) is defined identically to an MDP, except that the domains
 of $Tr$ and $R$ are finite sequences of states instead of single states:
$Tr: S^+ \times A \times S \rightarrow \Pi(S)$ and $R: S^+\times A \ \rightarrow \mathbb{R}$.
With this dependence on the history, to act optimally, a policy $\rho$ must, in general, take the form:
$ \rho : S^+ \rightarrow A $. Here $\rho$ is  a partial function defined on every sequence $h \in S^+$ reachable from $s_0$ under $\rho$. We define reachability under $\rho$ inductively: $s_0$ is reachable; if $h \in S^+ $ is reachable and $ Tr(h, a, s) > 0$ then $h \cdot s$ is reachable. 

The value of a trace $(s_0, s_1,..., s_n)$ is its discounted sum of rewards: $v(s_0 s_1,..,s_n) = \sum_{i=0}^n \gamma^n R(s_0,.., s_i)$. Because we assume the reward value is lower and upper bounded and $0<\gamma <1$, this discounted sum is always finite and bounded from above and below.

\subsection{RDPs}
A Regular Decision Processes (RDP)~\cite{BrafmanG19} is a factored NMDP in which the dependence on history is restricted
to regular functions.
%
 By {\em factored}, we mean that its states consist of assignments
to state variables and the transition function exploits this structure. Here, we assume Boolean state variables for convenience. To specify the dependence of the transition and reward on the history, we need a language that can compactly
specify sets of histories. Histories are essentially strings over an alphabet of states (for convenience, we can assume
that the last action executed is part of the state). Regular expressions (RE) are an intuitive and much used language for specifying languages, i.e., sets of strings. However, they do not have the logical structure to exploit more fine-grained properties of
the internal assignments of a state, and do not efficiently support various useful operations on sets of strings.
Linear dynamic logic on finite traces (\LDLf)~\cite{DeGVa13} combines linear-time temporal logic (LTL) with the syntax of propositional dynamic logic (PDL) but interpreted over finite traces. \LDLf has the same expressive power as RE, allows
us to refer to properties of states, and supports simple specification of conjunction, disjunction, and negation. 
RDPs use \LDLf formulas to specify properties or classes of histories. To simplify their exposition,
in this paper we do not make explicit use of them, and it is enough to know about RE and to think of each formula we mention as specifying an RE. For more details, see~\cite{BrafmanG19}.

An RDP is a tuple $M_L = \langle P, A, S, Tr_L, R_L, s_0 \rangle$. $P$ is a set of propositions inducing a state-space $S$ with $s_0$ as the initial state. $A$ is the set of actions. $Tr_L$ is a transition function represented by a finite set $T$ of quadruples of the form: $(\varphi, a, P', \pi(P'))$. $\varphi$ is an \LDLf formula over $P$, $a \in A$, $P' \subseteq P$ is the set of propositions affected by $a$ when $\varphi$ holds, and $\pi(P')$ is a joint-distribution over $P'$ describing its post-action distribution. The basic assumption is that the value of variables not in $P'$ is not impacted by $a$.

If $\{(\varphi_i, a, P'_i, \pi_i(P') | i \in I_a\}$ are all quadruples for $a$, then the $\varphi_i$'s must be mutually exclusive, i.e., $\varphi_i \wedge \varphi_j$ is inconsistent, for $i \neq j$. We also assume that the formulas are exhaustive. 

Letting $s|_{P'}$ denote $s'$ projected to $P'$, $Tr_L$ is defined as follows:
 $Tr_L((s_0,...,s_k), a, s') = \pi(s'|_{P'})$  if quadruple $(\varphi, a, P', \pi(P'))$ is the (single) one
    s.t. $s_1,...,s_k \models \varphi$ and $s_k$ and $s'$ agree on all variables in $P \setminus P'$.

That is, given current trace $s_0, ..., s_k$ and action $a$
let $(\varphi, a, P', \pi(P'))$ the quadruple with a condition $\varphi$ that is satisfied by $s_0, ..., s_k$ 
(by assumption, exactly one such $\varphi$ exists). 
$s'$ is a possible next state only if it assigns propositions
not in $P'$ exactly the same value to  as does $s_k$, i.e., they are not impacted by the action. Then, the probability that $s'$ is the next state
equals the probability $\pi$ assigns to the value of the $P'$ propositions in $s'$.

The reward function $R_L$ is specified via a finite set $R$ of pairs of the form $(\varphi, r)$. $\varphi$ is an \LDLf formula over $P$, and $r \in \mathbb{R}$ is a real-valued reward. Given a trace $s_0, ... , s_k$, the agent receives the reward: $R_L(s_0, ..., s_k) = \sum_{(\varphi, r) \in R, \, s0,...,s_k \models\varphi} r$. By definition $R_L$ is bounded above and below.

\subsection{Mealy-Machine}
A Mealy machine is a deterministic finite-state transducer whose output values are determined both by its current state and the current inputs. 
Formally, a Mealy Machine is a tuple $M = \langle S, s_0, \Sigma, \Lambda, T, G \rangle$. $S$ is the finite set of states, $s_0$ is the initial state. $\Sigma$ is the input alphabet and $\Lambda$ is the output alphabet. The transition function $T: S \times \Sigma \rightarrow S$ maps pairs of state and input symbol to the corresponding next state. The output function $G: S \times \Sigma \rightarrow \Lambda$ maps pairs of state and input symbol to the corresponding output symbol. 








\section{Representing and Solving RDPs}
Let $\varphi_1,\ldots,\varphi_n$ be the set of \LDLf formulas that specify the transitions and rewards of an RDP.
This set is finite because $T$ and $R$ are finite.
Since each formula is equivalent to an RE, there is an automaton that can track its satisfaction~\cite{GologTrans,DeGVa13}.
The automaton's input alphabet is the product of the sets of RDP states and actions, and it accepts a history IFF it 
satisfies the corresponding formula. Let $M_i$ be the automaton tracking $\varphi_i$. 
Let $M= \bigotimes M_i$ be the product automaton of all the $M_i$'s. 
This automaton will be at an accepting state of exactly one of its transition tracking components
given any string because the transition formulas are mutually exclusive and exhaustive.  In addition,
some reward tracking automata may also accept.

Building on the idea of using Mealy Machines to specify non-Markovian rewards~\cite{RewardMachines},
our key observation is that an RDP can also be specified by a Mealy Machine. The machine's input alphabet is 
the product of the sets of RDP states and actions. Its output function assigns to each triple of machine state $s_{Me}$, RDP state $s_{RDP}$
and action $a$, a set of propositions, $P_{s_{Me},(s_{RDP},a)} \subseteq P$, a distribution over their value in the next RDP state, and a reward.

Such a Mealy Machine represents an RDP because we can reconstruct the RDP from it as follows:
Given history $h$, let $s(h)$ be the Mealy Machine state reached on string $h$ from 
its initial state. Let $s_{RDP}$ be the current RDP state. The Mealy Machine output $G(h(s),(s_{RDP},a))$
provides us with a specification of the transition function and reward for history $h\cdot s_{RDP}$ and action $a$.

The Mealy Machine $M_{Me}$ that describes an RDP is constructed from
the product automaton $M$ defined above by adding to it an output function. Let
$s=(s_1,\ldots,s_k,\ldots,s_n)$ be a state of $M$, such that $M_k$ is the (only) transition tracking automaton in an accepting state.
That is $s_k$ is an accepting state of $M_k$. Let $\varphi_k$ be the formula which $M_k$ accepts, and let
$(\varphi_k, a, P', \pi(P'))$ be the corresponding quadruple. Let $r$ be the sum of rewards associated with
all the reward tracking automata that are in an accepting state in $s$. Define $G(s_{Me},(s_{RDP},a))=((P',\pi(P')),r)$, i.e., the propositions and
distributions associated with $\varphi_k$ and the sum of rewards of accepting reward-tracking automata.
%

The correctness of this construction follows from the definition of RDPs and the 
correctness of the construction of automata tracking the satisfiability of
\LDLf formulas.
The main benefit of this representation of an RDP is that it can serve as a target for learning algorithms that use existing methods
for learning Mealy Machines, as we do in Section 4.

To solve an RDP, we can transform it into an MDP $M_{MDP}$ by taking the product of
$M_{Me}$ with the original state space $S_{RDP}$ of the RDP~\cite{BrafmanG19}. 
The $M_{Me}$ state  reflects the relevant aspects of the entire history. 
Every $s_{RDP}\in S_{RDP}$ and $a\in A$ transform $M_{Me}$ deterministically,
but induce a Markovian stochastic transition over $S_{RDP}$.
$M_{MDP}$'s reward function is fully specified given the $M_{Me}$'s state, the current RDP state,
and the current action. 
$M_{MDP}$ can be solved using standard MDP solution techniques~\cite{Puterman}. 
We  use UCT~\cite{KocsisS06}, an MCTS algorithm, because MCTS can be applied to RDPs
without generating $M_{MDP}$ explicitly.
We maintain the current RDP state, i.e., the most recent set of observations, and the state $s_{Me}$ of the Mealy Machine.
From $s_{Me}$, for each action and current RDP state, we can obtain as output the information needed to sample the next
set of observations and rewards. The new observations replace the old ones, and are used (with the
action) to update the Mealy Machine. The choice of which action to apply follows the standard UCB1 criterion~\cite{AuerCF02}
$ a = arg\max_a Q(s_{Me}, a) + c\cdot \sqrt{\frac{\log{n(s_{Me})}}{n(s_{Me}, a)}}$, with $s_{Me}$ used here instead of the
current RDP state as it captures all history dependent properties of interest.

\section{Learning RDPs}
MDP learning algorithm rely on the Markov assumption and full observability of the state for their correctness. These algorithms
are not suitable for learning non-Markovian models, such as an RDP. 
Instead, we can exploit the alternative, Mealy Machine representation of RDPs and
use Mealy Machine learning algorithms to learn the RDP model.
More specifically, we use Flexfringe~\cite{VerwerH17} (using EDSM with the Mealy Machine heuristic) to
learn a Mealy Machine that represents the underlying RDP.
Then, we use MCTS to generate an optimal policy for the learned model. Thus, our approach can be characterized as a Model-based RL algorithm for non-Markovian domains.

\subsection{Learning Algorithm Overview}
Algorithm~1 provides the pseudo-code of our learning algorithm S3M (Sample, Merge, Mealy Machine). The next subsections describe each
step in  detail.

\begin{algorithm}[tb]
\caption{Sample Merge Mealy Model (S3M)}
\label{alg:algorithm1}
\textbf{Input}: $domain$\\
\textbf{Parameter}: $min\_samples$\\
\textbf{Output}: $M$
\begin{algorithmic}[1] 
\STATE Initialize state space of $M$ to RDP state space $S_{RDP}$
\REPEAT
\STATE Set $S=sample(domain)$.
\STATE Set $Tr=base\_distributions(S,min\_samples)$
\STATE $best\_loss = \infty$
\FOR{$\epsilon ~ in ~ possible\_epsilons$}
\STATE $Tr' = merger(Tr, \epsilon)$
\STATE $loss = calc\_loss(Tr', S)$
\IF{$loss < best\_loss$}
\STATE $Tr = Tr'$
\STATE $best\_loss = loss$
\ENDIF
\ENDFOR 
\STATE {\em Me = mealy\_generator}
\STATE Set $M= \langle P,A,S_{RDP} \times S_{Me}, Tr,R,(s_{0_M},s_{0_{Me}}) \rangle$
\UNTIL{Max\_Iterations}
\STATE \textbf{return} $M$
\end{algorithmic}
\end{algorithm}

A Mealy Machine learning algorithm expects input of the form {\em (input sequence, output)}, where {\em output} can be
the last output following this input sequence. Thus,
in our case, we need to generate inputs of the form  ($\pi,\alpha$), where $\pi$ is a trace
and $\alpha$ is a distribution over the next observation (RDP state) and reward. To create this input,
we first generate traces from the RDP by interacting with the environment (Line 3). These are traces of the
form $o_0,a_1,r_1,o_1,\ldots,a_k,r_k,o_k$, where $a_i$ is the action executed at the $i^{th}$ step,
$r_k$ is the reward received following its execution, and $o_i$ is the next RDP state. We use
$o_i$ to denote the RDP state to stress that it is fully observable.

To transform these traces to pairs of the form ($\pi,\alpha$),  we first identify a set of histories for which
we have enough samples ($\geq$ min\_samples). We refer to the empirical next-state distributions
associated with these histories as {\em base} distributions (L.~4). 
Next, in Lines 6-12, the rest of the histories are merged with the closest history based on the distance of their distribution from one of the base distributions. The merge choice depends on a parameter $\epsilon$, and we try a range of possible $\epsilon$ values, attempting to balance model size and accuracy.
Associating each of the resulting distribution with some symbol, we can now provide the needed input
to the Mealy Machine learning algorithm in the form of pairs {\em (trace,distribution)} (L.14). 
Finally, by taking the product of the learned Mealy Machine with the RDP's state space, we obtain an MDP that
can be solved for an optimal policy.

We repeat this process multiple times, each time with a more informed state space. Initially, we use the
RDP state space to guide exploration. Once we learn a Mealy machine, or improve our current
Machine,  we update the state space to reflect our new model to help better
guide  exploration.

\subsection{Sampling}
To learn the SDR model we need to generate sample traces. We considered two methods for doing this.
One is purely exploratory and does not attempt to exploit during the learning process, while the other
does. 

The {\em Pure Exploration} approach uses a stochastic policy that is biased towards actions that were sampled less.
More specifically, for every $a\in A$ and $s\in S_{RDP}$, where $S_{RDP}$ is the RDP state space, define:
\begin{align}
\label{E:sample}
   P(a|s) = \frac{f(a,s)}{\sum_a f(a,s)}\mbox{ where }f(a,s) = 1 - \frac{n(a,s)}{\sum_a n(a,s)} 
\end{align}
Here, $n(a,s)$ stands for the number of times action $a$ was performed in state $s$ of the RDP. This distribution favours actions that were sampled fewer times in a state. 

The {\em Smart Sampling} approach is essentially Q-learning~\cite{WatkinsD92} with some exploration using the above scheme. Specifically,
$Q$ values are maintained for each state-action pair, where states are defined and updated as above, starting with
a single-state Mealy Machine.  $Q(s,a)$ is initialize to 0 for all states and actions, and are updated following each sample
of the form $s,a,r,s'$
using $Q(s,a) = Q(s,a) + \alpha (R(s,a) +\gamma \max_{a'\in A} Q(s',a') - Q(s,a))$.  With probability $1-\epsilon$,
we select the greedy action in state $s$, and with probability $\epsilon$ we sample an action based on the
distribution defined in Equation~\ref{E:sample}.

\subsection{Trace Distributions}
Next, we associate with every trace encountered a set of propositions and a distribution over their probability.
(Note that each prefix of a trace is also a trace.)
Let $h = (o_1a_1,o_2a_2,...,o_{n},a_n)$ be a given trace. We define $P_{h,(o,a)}$, the set of propositions affected by
action $a$ given history $h\cdot o$, to be all propositions $p\in P$ such that there exists a trace 
$hoao' = (o_1a_1,o_2a_2,...,o_{n},a_n,o,a,o')$ in our sample where $o$ and $o'$ differ on the value of $p$.
We expect $P_{h,(o,a)}$ to be small, typically, because action effects are usually local.
Next, we compute the empirical post-action distribution over $P_{h,(o,a)}$ for history $h$, RDP state $o$ and action $a$. 
That is, the frequency of each assignment to $P_{h,(o,a)}$ in the last RDP state of over traces of the form $hoao'$ in our sample.


\subsection{Merging Histories and Their Distributions}
By modeling a domain as an RDP, our basic assumption is that what dictates the next state distribution of 
a history is the class of regular expressions it belongs to.
Hence, many different histories are likely to 
display similar behavior because they are instances of the same regular expression. 
Of course, we do not know what these regular expressions are,
and because of the noisy nature of our sample, we cannot expect two histories that belong to the same class
to have the same empirical distribution. Moreover, many histories will be sampled rarely, in which case their empirical next-state distribution is likely to significantly differ from the true one. For this reason, we attempt to cluster similar histories together based on their empirical
next-state distribution, using KL Divergence~\cite{KLD} as a distance measure. However, we consider merging only
histories that affect the {\em same} set of propositions.

Our goal is to create clusters s.t.~each cluster represents a certain distribution and each trace is assigned to a
single cluster.
We create the clusters bottom-up. First, we create a single cluster for each trace $hoa$ as described in 4.2. Each
such cluster has also a weight $w$ denoting the number of samples used to create it.
Then, for every two clusters with distributions $P_1$ and $P_2$ (affecting the same propositions) with weights $w_1\geq w_2 \geq
\mbox{min\_samples}$ , we merge them if:
\begin{enumerate}
\item The support of $P_2$ contains the support of $P_1$
\item $D_{KL}(P_1||P_2)\leq \epsilon$.
\end{enumerate}
Note that condition 1 is required for $D_{KL}$ to be well defined. 

The new cluster has weight $w=w_1+w_2$ and a distribution $P$ such that:
\begin{equation}
\label{E:merge}
P(\cdot) = (1/w) [w_1\cdot P_1(\cdot)  + w_2\cdot P_2(\cdot)]
\end{equation}
If a cluster has multiple other clusters with which it can merge, then the one with the smallest KL divergence is selected.
We repeat this procedure until no merges are possible.

Next, for each distribution $P$ whose weight $<$ min\_samples, we find the distribution
$Q$ from the above clusters that affects the same set of propositions
such that $D_{KL}(P||Q)$ is well defined and minimal, and merge the two using
Equation~\ref{E:merge} to obtain the new distribution. Notice that such a merge
implies that the support of $P$ is a subset of the support of $Q$.

The above is repeated for different values of $\epsilon$, resulting in different models. 
We now explain how we select the final model.
Each model has the form $(\Pi,Tr)$.
Each $\pi\in\Pi$ is a distribution over the set of assignments to some
subset $P_\pi$ of the RDP's set of propositions, with one such distribution associated
with each cluster. $Tr:H\rightarrow \Pi$ maps each history
in the sample to the distribution associated with its cluster.

To compare the models we define the following loss function:
\begin{align}
    loss =  -\sum_{h\in H} log(P(h |Tr(h)) + \lambda \cdot log(\sum_{\pi \in\ Pi}|supp(\pi)|) 
\end{align}
\begin{align}
    P(h |Tr(h)) = \prod_{i=1}^{n}P(o_i|o_1a_1 \ldots o_{i - 1}a_{i - 1};Tr(h))
\end{align}
Where $|supp(\pi)|$ is the size of the support  of distribution $\pi$. Thus, our loss function
is the log-likelihood of the data with a regulizer that penalizes models with many
clusters, and models with "mega"-clusters with large support. 

\subsection{Generate a Mealy Machine and an MDP}
We now use the {\em flexfringe} algorithm~\cite{VerwerH17} to learn a Mealy Machine from our data.
With every trace in our original sample, we associate the index of the cluster it belongs to.
The result is a Mealy Machine representing the RDP.

Let $M_{Me} = \langle S_{Me}, s_{0_Me}, \Sigma, \Lambda, T, G \rangle$ be the learned Mealy Machine.
From this Mealy Machine we can generate an MDP $M = \langle  S_{RDP}\times S_{Me}, A, Tr; R; (s_0,s_{0_{Me}})  \rangle$.
$S_{RDP}$ is the (observable) state space of the RDP; $A$ is the set of RDP actions; $s_0$ is the initial RDP state.
$Tr$ and $R$ are defined as follows: For a given state $(s_{RDP},s_{Me})$ of $M$ and action $a$, let
$G(s_{Me},(s_{RDP},a)) = ((P',\pi(P')), r)$. $Tr$ transforms the Mealy Machine state $s_{Me}$ deterministically based
on the transition function $T$ of  $M_{Me}$, i.e., to $T(s_{Me},(s_{RDP},a))$. It transforms the RDP state $s_{RDP}$ based on
the distribution $\pi(P')$ (leaving the value of all propositions in $P\setminus P'$ unchanged).
Finally, $R((s_{RDP},s_{Me}),a) = r$. The optimal policy for this MDP is optimal for the RDP -- associating actions
with regular functions of the RDP history.

\section{Empirical Evaluation}
As this is the first paper to explore the problem of learning and solving RDPs, our goal is to evaluate the
ability of our algorithms to address this problem, to assess our ability to scale-up, and to set up a baseline
for future work. To this effect, we define two new RDP domains and use them to test the two variants of our algorithm.

\subsection{The Domains}
We define two domains: Non Markovian Multi-Arm Bandit (MAB) and  Rotating Maze. 
The original MAB is stateless. In our version transition probabilities depend on the history of success for each arm, making it
a two states domain. The Maze problem is based on an agent navigating on a grid toward a designated location, while the orientation might change. 
For Mealy Machine learning we used the EDSM implementation from the FlexFringe library~\cite{VerwerH17}. To solve the learned RDPs
we use UCT, extended to RDPs,  and compare it with a baseline of $R_{\mbox{max}}$ -- a model-based MDP learning algorithm~\cite{BrafmanT02}. 
This learning algorithm essentially assumes (wrongly) that the RDP transitions are Markovian.

\subsubsection{Multi-Arm Bandit Domain}
The Multi-Arm Bandit (MAB) is the simplest class of RL domains. Standard MAB is state-less -- hence there
are no transition function to learn. At each step the agent chooses one of $n$ actions, and receives a reward that depends (possibly stochastically) on the choice of action. Our Non-Markovian MAB extends this by making the probability of
receiving a (fixed-size) reward depend on the entire history of previous actions. It is essentially a two-state RDP --
where the state indicates whether a reward was received or not. 
We created three  MAB-based RDP models:
\begin{enumerate}
    \item RotatingMAB: Let $\pi$ be a vector that assigns the probability of winning the reward for each action.
     This probability shifts right (i.e., $+1 \mbox{ mod } n$) every time the agent receives a reward. 
     Therefore, the probability to win for each arm depends on the entire history, but via a regular function.
    \item MalfunctionMAB: One of the arms is broken, s.t.~after the corresponding action is performed $k$ times,
    its probability of winning drops to zero for one turn.
    \item CheatMAB: There exists a sequence of actions s.t.~after performing that sequence, all actions lead to 
    a reward with probability 1 from that point on.

\end{enumerate}

In our experiment we used 2 arms/actions for all of the three variations of the domain. The winning probabilities of the machines of the RotatingMAB were $(0.9,0.2)$, for CheatMAB $(0.2,0.2)$ and for MalfunctionMAB $(0.8, 0.2)$.

\subsubsection{Maze Domain}
The Maze domain is  an $N \times N$ grid, where the agent starts in a fixed position. The possible actions of the agent are
{\em up/down/left/right}. These actions succeed $90\%$ of the time, and in the rest $10\%$ the effect is to
move in the opposite direction. The goal of the domain is to get to the designated location where a final reward is received. In a normal MDP this task would be quite easy to solve using conventional RL algorithms. However, in this maze, every three actions the agent's orientation changes by $90^{\circ}$.
Thus, the effects of the actions are a regular function of the history.
In our experiment we used a $4\times4$ maze, where the goal is five steps away from the initial position.


%
%
%
%
%

\begin{figure*}[bt]
\centering     
\subfigure[Rotating MAB Domain Results]{\label{fig:a}\includegraphics[width=0.97\columnwidth]{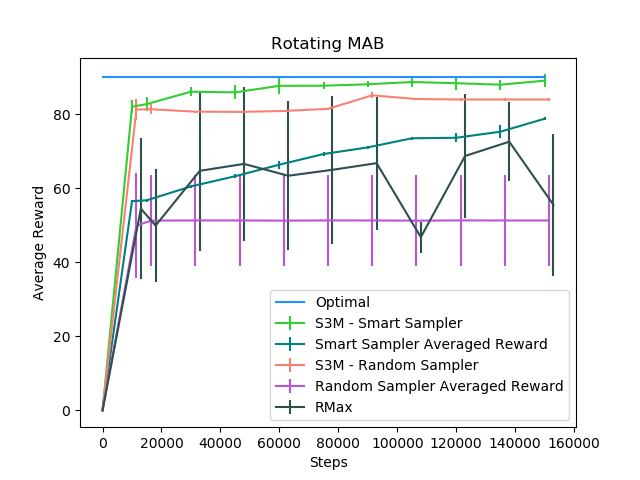}
\label{sub-fig-rot-mal-1}
}
\subfigure[Malfunction MAB Domain Results]{\label{fig:b}\includegraphics[width=0.97\columnwidth]{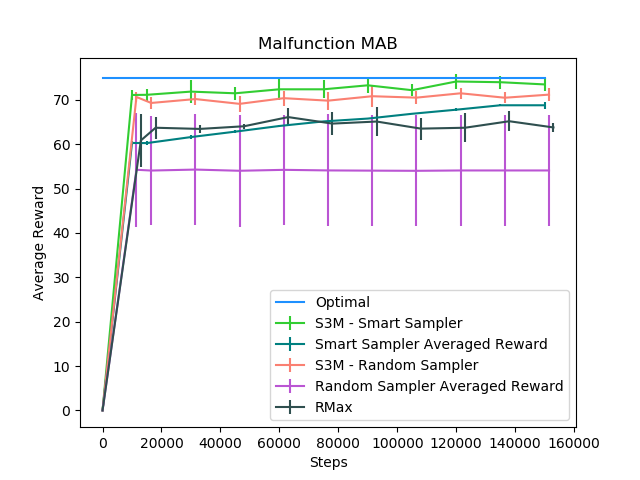}
\label{sub-fig-rot-mal-2}
}

%
\subfigure[Cheat MAB Domain Results]{\label{fig:a}\includegraphics[width=0.97\columnwidth]{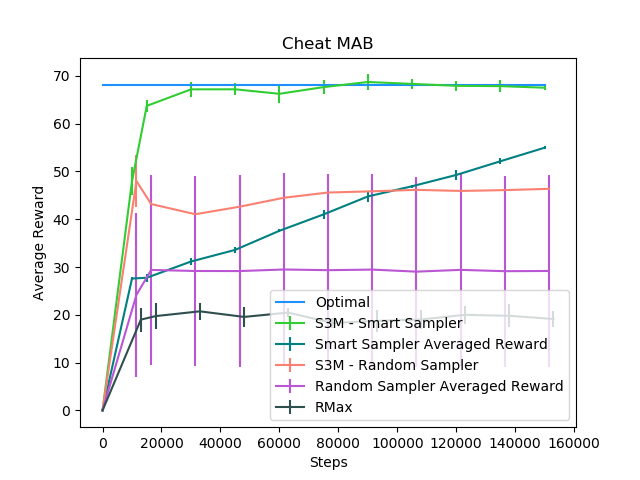}
\label{sub-fig-maze-cheat-1}
}
\subfigure[Maze Domain Results]{\label{fig:b}\includegraphics[width=0.97\columnwidth]{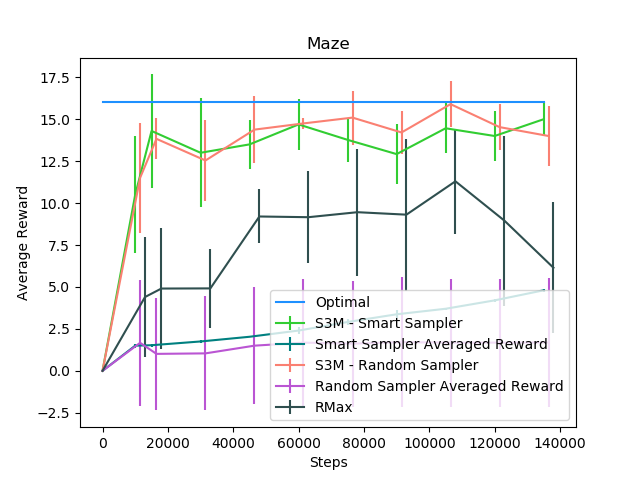}
\label{sub-fig-maze-cheat-2}
}
\caption{Results}
\label{avg-reward}
\end{figure*}


\subsection{Results}

For each domain we used three configurations: (1) {\em Random Sampler:} S3M with {\em Pure Exploration} sampling;
(2) {\em Smart Sampler:} S3M with {\em Smart Sampling}; 
(3) {\em RMax:} The model-based RL algorithm $R_{Max}$~\cite{BrafmanT02} that uses the RDP states as its states, 
used as baseline. 

The results are displayed in Figure~\ref{avg-reward}. We show the value of the optimal policy ({\em Optimal}), the quality
of the policies learned by the above three algorithms at each evaluation step, and the average reward accumulated during
learning by the first two algorithms.  Each experiment was repeated five times. We plot the average over these five repetitions
with error bars representing the std. To evaluate the quality of the current policy during the learning process, 
the optimal policy for the current model was computed online using UCT.
50 trials were conducted using the currently learned Mealy Machine.  MAB trials were 10 steps long, and Maze trials were terminated after 15 steps if the goal was not reached. The graph shows the average (over 50 trials) per-step rewards of these policies (averaged over 5 trials),
and the average  (over 5 trials) accumulated reward  obtained as S3M was sampling traces.



Overall, S3M Smart Sampler does quite well, yielding optimal or near optimal average reward. It also does well in terms of
accumulated rewards. Random Sampler does reasonably well, too, except on the CheatMAB, but its accumulate
reward is typically much worse. $R_{max}$, which cannot capture the non-Markovian dynamics does much worse.
Even in the Maze domain, a closer look reveals that the difference in its performance
is exactly what we would expect when we ignore the non-Markovian behavior -- this margin in that case is simply smaller than in
the MAB domains.

Proper exploration is a key issue in RL. In the case of RDPs, it is not enough to explore states because we need to gather statistics on histories. In this respect, it is interesting to compare the performance of the two sampling approaches. We expected that 
Smart Sampler will accumulate more reward, as it does more exploitation and this exploitation is informed by the learned Mealy Machine, 
and hence takes history into account. Indeed, in the MAB domain, we see such behavior:
S3M with smart sampling converges to an optimal or near-optimal policy quickly, and the accumulated rewards 
increase steadily, while the random sampler does worse. This is especially pronounced in the Cheat MAB domain, which is the most
complex domain. We believe the reason for the weak performance of random sampling in this domain is that too many of the samples
are not along the more interesting traces that discover the "cheat" sequence. Therefore, it is more difficult for it to learn a good Mealy
Machine that can exploit it.  Surprisingly, in the Maze domain, unlike in the MAB domains, there is no significant difference between the two S3M versions. We hypothesize that in Maze, there is more to learn about the general behavior of transitions because there are more states, and  the random sampler generates more diverse samples that provide a more accurate statistics on various states. The smart sampler, on the other hand, does not. Moreover, while the domain has more states, the regular expression that governs the dynamics is relatively simple, and so the random sampler is still able to learn the corresponding automaton. 

Generally speaking, we see a high correlation between the quality of the samples collected by the sampler and the quality of the learned model: when the average reward of the samples is monotonically increasing so does the averaged reward of the policy obtained by MCTS. An open question is what is the exact relation: do samples that concentrate along desirable traces yield better Mealy Machines, or is it the case that because we learn a better Mealy Machine, the traces generated by the Smart Sampler have higher rewards (naturally).

\section{Discussion and Summary}
We presented the first algorithm for learning Regular Decision Processes. By viewing the RDP specification as a Mealy Machine, we were able to combine Mealy Machine and RL algorithms to obtain an algorithm for learning RDPs that quickly learns a good
Mealy Machine representation in our experiments. Naturally, there is much room for improvement, especially in methods for better sampling and better aggregation of histories.

\bibliographystyle{named}
\bibliography{ijcai20}

\begin{thebibliography}{}

\bibitem[\protect\citeauthoryear{Auer \bgroup \em et al.\egroup
  }{2002}]{AuerCF02}
Peter Auer, Nicol{\`{o}} Cesa{-}Bianchi, and Paul Fischer.
\newblock Finite-time analysis of the multiarmed bandit problem.
\newblock {\em Machine Learning}, 47(2-3):235--256, 2002.

\bibitem[\protect\citeauthoryear{Baier \bgroup \em et al.\egroup
  }{2008}]{GologTrans}
Jorge~A. Baier, Christian Fritz, Meghyn Bienvenu, and Sheila McIlraith.
\newblock Beyond classical planning: Procedural control knowledge and
  preferences in state-of-the-art planners.
\newblock In {\em Proceedings of the 23rd AAAI Conference on Artificial
  Intelligence (AAAI), Nectar Track}, pages 1509--1512, Chicago, Illinois, USA,
  July 13-17 2008.

\bibitem[\protect\citeauthoryear{Brafman and {De Giacomo}}{2019}]{BrafmanG19}
Ronen~I. Brafman and Giuseppe {De Giacomo}.
\newblock Planning for ltlf /ldlf goals in non-markovian fully observable
  nondeterministic domains.
\newblock In {\em Proceedings of the Twenty-Eighth International Joint
  Conference on Artificial Intelligence, {IJCAI} 2019, Macao, China, August
  10-16, 2019}, pages 1602--1608, 2019.

\bibitem[\protect\citeauthoryear{Brafman and Tennenholtz}{2002}]{BrafmanT02}
Ronen~I. Brafman and Moshe Tennenholtz.
\newblock {R-MAX} - {A} general polynomial time algorithm for near-optimal
  reinforcement learning.
\newblock {\em J. Mach. Learn. Res.}, 3:213--231, 2002.

\bibitem[\protect\citeauthoryear{Camacho \bgroup \em et al.\egroup
  }{2019}]{RewardMachines}
Alberto Camacho, Rodrigo~Toro Icarte, Toryn~Q. Klassen, Richard Valenzano, and
  Sheila~A. McIlraith.
\newblock {LTL} and beyond: Formal languages for reward function specification
  in reinforcement learning.
\newblock In {\em Proceedings of the Twenty-Eighth International Joint
  Conference on Artificial Intelligence ({IJCAI})}, 2019.

\bibitem[\protect\citeauthoryear{De~Giacomo and Vardi}{2013}]{DeGVa13}
Giuseppe De~Giacomo and Moshe~Y. Vardi.
\newblock Linear temporal logic and linear dynamic logic on finite traces.
\newblock In {\em IJCAI-13}, pages 854--860, 2013.

\bibitem[\protect\citeauthoryear{Kocsis and Szepesv{\'{a}}ri}{2006}]{KocsisS06}
Levente Kocsis and Csaba Szepesv{\'{a}}ri.
\newblock Bandit based monte-carlo planning.
\newblock In {\em Machine Learning: {ECML} 2006, 17th European Conference on
  Machine Learning, Berlin, Germany, September 18-22, 2006, Proceedings}, pages
  282--293, 2006.

\bibitem[\protect\citeauthoryear{Kullback and Leibler}{1951}]{KLD}
Solomon Kullback and Richard Leibler.
\newblock On information and sufficiency.
\newblock {\em Annals of Mathematical Statistics}, 22(1):79--86, 1951.

\bibitem[\protect\citeauthoryear{Puterman}{2005}]{Puterman}
Martin~L. Puterman.
\newblock {\em Markov Decision Processes: Discrete Stochastic Dynamic
  Programming}.
\newblock Wiley, 2005.

\bibitem[\protect\citeauthoryear{Silver and Veness}{2010}]{SilverV10}
David Silver and Joel Veness.
\newblock Monte-carlo planning in large pomdps.
\newblock In {\em Advances in Neural Information Processing Systems 23: 24th
  Annual Conference on Neural Information Processing Systems 2010. Proceedings
  of a meeting held 6-9 December 2010, Vancouver, British Columbia, Canada},
  pages 2164--2172, 2010.

\bibitem[\protect\citeauthoryear{Sutton and Barto}{1998}]{Sutton2}
R.~S. Sutton and A.~G. Barto.
\newblock {\em Reinforcement Learning: An Introduction}.
\newblock MIT Press, Cambridge, MA, 1998.

\bibitem[\protect\citeauthoryear{Verwer and Hammerschmidt}{2017}]{VerwerH17}
Sicco Verwer and Christian~A. Hammerschmidt.
\newblock flexfringe: {A} passive automaton learning package.
\newblock In {\em 2017 {IEEE} International Conference on Software Maintenance
  and Evolution, {ICSME} 2017, Shanghai, China, September 17-22, 2017}, pages
  638--642, 2017.

\bibitem[\protect\citeauthoryear{Watkins and Dayan}{1992}]{WatkinsD92}
Christopher J. C.~H. Watkins and Peter Dayan.
\newblock Technical note q-learning.
\newblock {\em Machine Learning}, 8:279--292, 1992.

\end{thebibliography}

\end{document}